\documentclass{article}

\usepackage{microtype}
\usepackage{graphicx}
\usepackage{subfigure}
\usepackage{booktabs}
\usepackage{amsmath,amssymb}
\usepackage{gensymb}
\usepackage{url}
\usepackage{stfloats}

\usepackage{hyperref}

\usepackage{xspace}
\newcommand{\am}{\textsc{AugMix}\xspace}
\usepackage[accepted]{icmlish}

\icmltitlerunning{Diverse Ensembles Improve Calibration}

\begin{document}

\twocolumn[
\icmltitle{Diverse Ensembles Improve Calibration}

\icmlsetsymbol{equal}{*}

\begin{icmlauthorlist}

\icmlauthor{Asa Cooper Stickland}{ed}
\icmlauthor{Iain Murray}{ed}

\end{icmlauthorlist}

\icmlaffiliation{ed}{School of Informatics, University of Edinburgh}

\icmlcorrespondingauthor{Asa Cooper Stickland}{a.cooper.stickland@ed.ac.uk}

\icmlkeywords{Robust Machine Learning, Uncertainty, Ensembles, Data Augmentation}

\vskip 0.3in
]

\printAffiliationsAndNotice{} 
\begin{abstract}
Modern deep neural networks can produce badly calibrated predictions,
especially
when train and test distributions are mismatched. Training an ensemble of models and averaging their predictions can help alleviate these issues. We propose a simple technique to improve calibration, using a different data augmentation for each ensemble member. We additionally use the idea of `mixing' un-augmented and augmented inputs to improve calibration when test and training distributions are the same. These simple techniques improve calibration and accuracy over strong baselines on
the CIFAR10 and CIFAR100 benchmarks, and out-of-domain data from their corrupted versions.
\end{abstract}

\section{Introduction}
\label{sec:intro}

Modern neural network models can produce overconfident or miscalibrated predictions, even when training examples are independent and identically distributed (i.i.d.)\ to the test distribution. This miscalibration is exacerbated when the training and testing distributions are different. In safety-critical scenarios, the ability to accurately represent model uncertainty is valuable.  

Such model miscalibration has been shown to reduce when we train an ensemble of models and average their predictions \cite{NIPS2017_7219,NIPS2019_9547}.
Ensembles have long been known to improve generalisation \cite{58871}, especially when an ensemble is diverse, which is promoted with various techniques such as using latent variables \cite{sinha2020dibs} or diversity-encouraging losses and architecture changes \cite{Kim_2018_ECCV, NIPS2016_6270, pmlr-v97-pang19a}. 
Recent work proposes `cheap' ensembles by sharing most of the model parameters across all ensembles, and using rank-1 factors to modify the linear layers in an ensemble member \cite{Wen2020BatchEnsemble:}, making ensembles easier to train and store.

Another long-standing way to improve generalization is \textit{Data Augmentation}, i.e.\ expanding our training set with modified copies \cite{zhang2018mixup, Yun2019CutMixRS, Cubuk_2019_CVPR}. Recent examples include work by \citet{hendrycks*2020augmix} and \citet{DBLP:journals/corr/abs-1911-04252}.
These approaches exploit the intuition that a blurry or rotated image should have the same class as the original image.

This work extends and combines recent work on cheap ensembles and data augmentation.
We increase ensemble diversity by applying different augmentations to each ensemble member.
This method improves calibration on an i.i.d.\ test set, and accuracy and calibration on out-of-distribution test sets for the CIFAR10 and CIFAR100 datasets. We additionally simplify the idea of `mixing' un-augmented and augmented inputs introduced by \citet{hendrycks*2020augmix}, and explore adversarial perturbations, which apply more generally and result in better performance on i.i.d.\ data.

\section{Methods}
\label{sec:methods}
The computational and space costs of training independent ensembles scale linearly with the number of ensembles, so we share parameters as in BatchEnsemble \cite{Wen2020BatchEnsemble:}.
Each weight matrix in ensemble member $i$ is a Hadamard product $W_i = W \circ r_i s_i^\top$, where $W$ is shared across all ensemble members, and $r_i$ and $s_i$ are vectors that adapt the weights for this member. The adaptation can be implemented efficiently by elementwise multiplication of hidden states by $s_i$ before multiplication by $W$, and $r_i$ after.

We take a batch of $B$ training examples, repeat it $K$ times, where $K$ is the number of ensemble members, and apply different augmentations to each copy of the original batch. When we pass the examples through the network, each copy of an example sees a different set of adapted weights.

We aim to test if different augmentations in each ensemble member gives better calibration, we use three augmentations:
    1)~\textit{Adversarial perturbations:}
        Images are perturbed to increase training loss, adapting the `fast gradient sign method' \citep{goodfellow2014explaining}.
    2)~\textsc{AugMix}: Augmentations from \citet{hendrycks*2020augmix}, with minor modifications.
    3)~\textit{Stochastic Depth} \citep{Huang2016DeepNW, Fan2020Reducing}. We randomly drop residual connections.
With the exception of Stochastic Depth, we can apply these augmentations to the input data and require no modification of the neural network architecture.

\subsection{Adversarial Perturbations}
\label{sec:adv}

We generate a new input image $x_{\mathrm{adv}}$, from the original \{image, label\} pair $\{x, y\}$ as follows:
\begin{equation}
\label{eq:per}
    x_{\mathrm{adv}} = x + \frac{m}{\mathbb{E}(m)} \cdot u \cdot s \cdot \mathrm{sign}(\nabla L(x, y)),
\end{equation}
where $L(x, y)$ is the training loss, $u \sim U(0, 1)$, $m \sim \mathrm{Bernoulli(p)}$ and $s$ is a constant, the `severity' of augmentation that varies per ensemble member. Without $u$ and $m$, it's the \emph{fast gradient sign method} \citep{goodfellow2014explaining}. 

We introduced $u$ to create a distribution of perturbations, and $m$ so that we sometimes get an unperturbed image as input. Preliminary experiments suggested that these additional terms improved performance.
The random $m$ and $u$ are scalars, i.e.\ they vary by input example but not by input dimension. The $m$ term, scaled by its expectation, is taken from Dropout \cite{srivastava2014}.

\subsection{\am}
\label{sec:am}

The \am method
\citep{hendrycks*2020augmix}
aims to make models robust to out-of-distribution data
by exposing the model
to a wide variety of
augmented images.
The augmentation operations are from AutoAugment \cite{Cubuk_2019_CVPR}, excluding the operations used to create out-of-domain corrupted test sets (Section~\ref{sec:data}). Most operations have a varying `severity', e.g.\ rotation by 2\degree~or 15\degree. Several augmentation `chains' are sampled, where a chain is a composition of one to three randomly selected operations. The augmented images from each chain are combined
with a random convex combination, see the `Augment' function in Algorithm~\ref{alg:example}.

The final stage of \am combines the original and augmented image with a convex combination sampled from $\mathrm{Beta}(\alpha, \alpha)$.
Initial results of \citet{hendrycks*2020augmix} suggest a bimodal ($\alpha=0.1$) Beta distribution performed best\,---\,sometimes using an image close to the original one.
We consider a simpler approach: simply picking the original or augmented image using a Bernoulli trial, as we did in Section~\ref{sec:adv}. To encourage diversity, we additionally use a different augmentation severity per ensemble member. Our modified \am procedure is described in Algorithm~\ref{alg:example}.

\begin{algorithm}[tb]
    \small
   \caption{Modified \am Pseudocode}
   \label{alg:example}
\begin{algorithmic}
   \STATE {\bfseries Input:} Image $x_{\mathrm{orig}}$, Severity Vector $\mathbf{s}$, Ensemble Index $i$
   \FUNCTION {Augment($x_\mathrm{orig}$,\, $s$, $\,k\!=\!3,\,\alpha\!=\!1$)}
		\STATE Fill $x_\mathrm{aug}$ with zeros
		\STATE Sample mixing weights:\\ \quad$(w_1, w_2, \ldots, w_k) \sim \text{Dirichlet}(\alpha,\alpha,\ldots,\alpha)$
		\FOR{$i = 1, \ldots, k$}
        \STATE Sample aug.\ operations $\text{op}_1, \text{op}_2, \text{op}_3 \sim \mathcal{O}$
        \STATE Compose operations with varying depth (with severity $s$) $\text{op}_{12} = \text{op}_2 \circ \text{op}_1$ and  $\text{op}_{123} = \text{op}_3 \circ \text{op}_2 \circ \text{op}_1$
        \STATE Sample uniformly $\text{chain} \sim \{\text{op}_1, \text{op}_{12}, \text{op}_{123} \}$
        \STATE $x_\mathrm{aug} \mathrel{+}= w_i \cdot \text{chain}(x_\mathrm{orig})$
		\ENDFOR
	\STATE {\bfseries return} {$x_{\mathrm{aug}}$}
	\ENDFUNCTION
   \FUNCTION {AugmentAndMix($x_{\mathrm{orig}},\, s,\, p\!=\!0.875, \beta\!=\!1.0$)}
   \STATE {$x_{\mathrm{aug}} = \mathrm{Augment}(x_{\mathrm{orig}}, s)$ }
   \STATE {Sample weight $m \sim \text{Bernoulli(p)}$ ~or~ $\text{Beta}(\beta,\beta)$}
   \STATE {Interpolate $x_{\mathrm{augmix}} = (1-m) x_{\mathrm{orig}} + m x_{\mathrm{aug}}$}
   \STATE {\bfseries return} {$x_{\mathrm{augmix}}$}
   
   \ENDFUNCTION
   \STATE {$x_{\mathrm{augmix}}^i=\mathrm{AugmentAndMix}(x_{\mathrm{orig}},\, s\!=\!\mathbf{s}_i)$}
\end{algorithmic}
\end{algorithm}

The full \am method encourages consistency across predictions for diverse augmentations of the same input, through the use of the Jensen--Shannon divergence as a consistency loss. Note this consistency loss can be applied to one model, and could be used for each ensemble member separately (i.e.\ it is not intended to ensure members agree). However due to the extra complication of using the consistency loss, and the doubling or tripling of the batch size to generate multiple samples of the same batch (on top of increasing the batch size for the BatchEnsemble method), we did not use this aspect of the \am method.

\begin{table*}[t]
\vspace*{-0.1in}
\caption{Comparing performance when using the \am  (referred to as `AM' in the table) method, with either a Beta distribution to mix augmented images, or a Bernoulli (`Bern.') distribution, with $p$ being the probability of augmenting an input image. `No B.E.' refers to using a single model, and otherwise we use BatchEnsemble. 
We highlight in bold the best result for each metric, separately for the test set.} 
\label{tab:aug}
\vskip 0.15in
\begin{center}
\scalebox{0.9}{
\begin{small}
\begin{sc}
\begin{tabular}{lccccccr}
\toprule
 & Cifar 10 &  & & Cifar 100 & & & \\

Method & Val.\ err.  & Val. ECE  & Val. ECE-rms & Val.\ err.  & Val. ECE  & Val. ECE-rms \\
\midrule
AM Beta & 3.82 & 1.41 & 3.60 &{\bf 18.5} & 4.69 & 5.6\\
AM Beta (not diverse)     & 3.88 & 1.48 & 3.19 &{\bf 18.5} & 4.76 & {\bf 5.46}\\
AM Bern.\ ($p=0.875$)  & 3.46 & 1.26 & 2.63 & 18.6 &{\bf 4.47} & 5.58 \\
AM Bern.\ ($p=1.0$)    & 3.64 & 1.38 & 3.28 & 18.8 & 5.07 & 5.90 \\
\midrule
AM Bern.\ ($p=0.875$) + Adv. &{\bf 3.44 }  &{\bf 1.15 }&{\bf 1.99} & 19.0 & 5.21 & 6.11 \\
\midrule
\midrule
 & Test &  & & Test & & & \\
\midrule
 & Err.  & ECE  & ECE-rms & Err.  & ECE  & ECE-rms \\
 AM (no B.E.) Bern.\ ($p=0.875$)  & 3.98 & 1.15 & 2.13 & 21.0 & 7.41 & 9.37 \\
 AM Bern.\ ($p=0.875$)  & 3.40 & 1.15 & 2.04 &  17.7 & 4.95 & 6.08 \\
AM Bern.\ ($p=0.875$) + Adv. & {\bf 3.13} & {\bf 1.00} & {\bf 1.88} & {\bf 17.6} & {\bf 4.51} & {\bf 5.36} \\
\midrule
\midrule

 & Cifar 10-C &  & & Cifar 100-C & & & \\

 & Err.  & ECE  & ECE-rms & Err.  & ECE  & ECE-rms \\

\midrule
AM Beta   & 11.6 & {\bf 4.30} & {\bf 5.61} &  34.0 & 10.2 & 11.0\\
AM Beta (not diverse)   & 11.4 & 4.39 & 5.88 & 34.3 & 10.7 & 11.7\\
AM Bern.\ ($p=0.875$)  & 11.5 & 4.78 & 6.147 & 34.0 & {\bf  9.92} &  {\bf 10.8} \\
AM Bern.\ ($p=1.0$)    & {\bf 11.3} & 4.63 & 5.99 &{\bf 33.7} & 10.7 & 11.7 \\
\midrule
AM Bern.\ ($p=0.875$) + Adv. & 11.6 & 4.72 & 6.07 & 34.2 & 10.8 & 11.8 \\
\midrule
\midrule
 & Test &  & & Test & & & \\
\midrule

 & Err.  & ECE  & ECE-rms & Err.  & ECE  & ECE-rms \\
 AM (no B.E.) Bern.\ ($p=0.875$)  & 12.8 & 5.27 & 6.93 & 35.8 & 14.8 & 16.6 \\
 AM Bern.\ ($p=0.875$)  & 11.0 & 4.32 & 5.71 & {\bf 33.2} &  10.4 & 11.5 \\
AM Bern.\ ($p=0.875$) + Adv. & {\bf 10.6} & {\bf 4.08} & {\bf 5.40} & {\bf 33.2} & {\bf 10.2} & {\bf 11.2} \\

\bottomrule
\end{tabular}
\end{sc}
\end{small}}
\end{center}
\vskip -0.1in
\end{table*}

\section{Experimental Settings}
\vspace*{-0.01in}

\subsection{Datasets}
\vspace*{-0.01in}
\label{sec:data}
We use the CIFAR-10 and CIFAR-100 classification datasets of tiny natural images \cite{cifar}.
For each task, we hold out a validation set of 5,000 random images from the training set, and train the best-performing models on the entire training set and evaluate on the test set. To test robustness to out-of-distribution data, we evaluate on corrupted versions of the test sets, CIFAR-10-C and CIFAR-100-C \cite{hendrycks2019robustness}. There are 15 noise, blur, weather, and digital corruption types, each appearing at 5 severity levels or intensities. We do not use these corruptions in training.

\subsection{Metrics}
 
When a well calibrated model is, say, 60\% confident, it will be correct 60\% of the time. A measure of this is \emph{Expected Calibration Error} \citep[ECE;][]{pmlr-v70-guo17a}. We divide the data into $m$ equally sized bins $B_m$ and measures the absolute difference between average confidence and accuracy in each bin, i.e.\ $\mathrm{ECE}=\sum_{m=1}^M \frac{|B_m|}{n}|\mathrm{acc}(B_m) - \mathrm{conf}(B_m)|$, where $n$ is the number of examples. An alternative \cite{hendrycks*2020augmix} uses squared difference instead of absolute, and takes the square root of the sum, which we call `ECE-rms'. We always use 15 bins. For corrupted data we sum error over the corruption intensities, and average over the 15 corruption types~\cite{hendrycks*2020augmix}.

\subsection{Training Setup}

We use the ResNeXt-29 ($32{\times}4$) architecture \cite{resnext}, which was the best-performing of those tested by \citet{hendrycks*2020augmix}. We use SGD with Nesterov momentum, an initial learning rate of 0.1 decayed following a cosine schedule \cite{DBLP:conf/iclr/LoshchilovH17}, and weight decay of 0.0005. Input images are pre-processed with standard random left-right flipping and cropping prior to any augmentations. We train ensembles for 250 epochs, and otherwise train for 200 epochs, following \citet{Wen2020BatchEnsemble:}.
We initialise the per-ensemble parameters (the vectors that produce the rank-1 modification of the weight matrices) for BatchEnsemble with a $N(1,0.5^2)$ distribution, and always use 4 ensemble members. We use a batch size of 128, which is 4$\times$ larger when using BatchEnsemble (each ensemble member sees a copy of the same 128 images). 

We introduce a `severity vector' $\mathbf{s}$, where each element of the vector corresponds to  the severity of augmentation for a particular ensemble member. By varying this vector we can control whether each ensemble member gets the same input distribution, or a different one per ensemble member.
Our augmentation hyperparameters are as follows:
\begin{itemize}
    \item \emph{Adversarial perturbations:} $\mathbf{s} = [0.0, 0.05, 0.1, 0.15]$, with $s_i$ corresponding to $s$ in eq~\ref{eq:per}. The probability of not using an augmented image was $p\!=\!0.875$. When not using different severity per ensemble member, we randomly shuffle the elements of $\mathbf{s}$ before each update.
    \item \am: $\mathbf{s} = [1, 2, 3, 4]$ (severity in \am takes on integer values). When not using different severity per ensemble member, we set all elements to 3, the default value used by \citet{hendrycks*2020augmix}.
    \item \emph{Stochastic Depth:} $\mathbf{s} = [0.0, 0.05, 0.1, 0.15]$, with the elements corresponding to probability of dropping a residual connection. When not using different severity per ensemble member, we set all elements to 0.075.
\end{itemize}

\begin{table*}[t]
\vspace*{-0.1in}
\caption{Accuracy and calibration for various methods on the cifar10 and cifar100 validation set (I.I.D. to the training distribution). `B.E.' refers to `BatchEnsemble'. `$p=1.0$' refers to always using the perturbed input, rather than skipping with some probability as in eq~\ref{eq:per}. `S.D.' refers to stochastic depth (randomly dropping residual connections). We highlight in bold the best result for each metric.}
\label{tab:iid}
\vskip -0.15in
\begin{center}
\scalebox{0.9}{
\begin{small}
\begin{sc}
\begin{tabular}{lccccccr}
\toprule
 & Cifar 10 &  & & Cifar 100 & & & \\

Method & Val.~err.  & Val. ECE  & Val. ECE-rms & Val.~err.  & Val.~ECE  & Val.~ECE-rms \\

Vanilla ResNeXt    & 4.18 &{\bf  1.25}& 3.20 & 21.0 & 6.88 & 8.06 \\
+ BatchEnsemble & 3.96 & 1.57& 3.07 & 19.4 & 5.04 & 5.81\\
\midrule
B.E. + S.D. &{\bf 3.30} & 1.45 & 2.76 & 19.6 & 8.50 & 10.2\\
B.E. + S.D. (not diverse)       & 3.60 & 1.70 & 3.85 & 19.9 & 9.91 & 11.7 \\

B.E. + Adversary & 3.60 & 1.29 & {\bf 2.36} & 19.5 & 4.75 & 5.57\\
B.E. + Adversary (not diverse)       &  3.46 & 1.34 & 2.64 &{\bf 19.1} &{\bf 4.49} & {\bf 5.31}\\
B.E. + Adversary ($p=1.0$)     & 4.12 & 1.56 & 3.58 & 19.5 & 4.92 & 6.13\\

\midrule
\midrule

 & Cifar 10-C &  & & Cifar 100-C & & & \\

 & Err.  & ECE  & ECE-rms & Err.  & ECE  & ECE-rms \\

Vanilla ResNeXt    & 26.5 & 13.5 & 15.4 & 50.1 & 19.9 & 22.1 \\
+ BatchEnsemble  & 27.0 & {\bf 13.2} & {\bf 15.0} & 50.8 & 19.9 & 21.1\\
\midrule
B.E. + S.D. & 26.3 & 14.2 & 16.2 &{\bf  50.0} & 24.7 & 26.2\\
B.E. + S.D. (not diverse)       & {\bf 26.0} & 15.6 & 17.6 & 50.2 & 28.7 & 30.3 \\

B.E. + Adversary & 26.2 &{\bf  13.2} & {\bf 15.0} & 50.6 &{\bf  18.9} & {\bf 19.9}\\
B.E. + Adversary    (not diverse)   & 26.5 & 14.3 & 16.1 & 50.8 & 19.9 & 21.0\\
B.E. +  Adversary ($p=1.0$)     & 26.6 & 13.3 & 16.1 & 50.6 & 20.3  & 21.5 \\

\bottomrule
\end{tabular}
\end{sc}
\end{small}}
\end{center}
\vskip -0.1in
\end{table*}

\section{Experiments and Discussion}

We group our results by those with the \am method (Table~\ref{tab:aug}) and those without it (Table~\ref{tab:iid}). Baselines (without augmentation) with single models and BatchEnsemble are at the top of Table~\ref{tab:iid}.

\subsection{Diverse Inputs vs. Not Diverse Inputs}

For each of the augmentation types we considered, \am, adversarial perturbations and stochastic depth, using a different `severity' of augmentation for each ensemble member tended to improve calibration on both i.i.d.\ and out-of-domain held-out data. Comparing methods to their `not diverse' counterpart in Table~\ref{tab:aug} and Table~\ref{tab:iid}, the ECE and ECE-rms scores for `not diverse' ensembles are generally lower. This effect did not always hold for i.i.d.\ held out data, but it was always true for CIFAR10-C and CIFAR100-C\@. The error rate, however, showed no clear trend between the two types of ensemble.

Dropping residual connections (`stochastic depth') results in slightly better performance on CIFAR10, but decreases calibration performance on CIFAR10-C and CIFAR100-C. 

\subsection{The Importance Of Un-augmented Inputs}

For the \am and adversarial perturbation methods, we compared two settings: augmenting every input image ($p=1.0$ in Tables~\ref{tab:aug} and~\ref{tab:iid}), or randomly mixing in some un-augmented images with 12.5\% probability ($p=0.875$ in Tables~\ref{tab:aug} and~\ref{tab:iid}).

We find (Table~\ref{tab:aug} and Table~\ref{tab:iid}), when using ensembles, that augmenting all inputs (with either \am or an adversary) results in worse performance on i.i.d.\ data, in terms of accuracy and calibration. For corrupted data the picture is mixed: for adversarial perturbation, augmenting all inputs gives similar or worse calibration, but for \am accuracy improves on corrupted images when augmenting all images (and calibration for CIFAR10-C)\@. We also compare, in Appendix~\ref{app}, augmenting every input image and mixing in un-augmented images for single models (i.e.\ not BatchEnsemble) for \am (Table~\ref{tab:app-aug}), however no clear picture emerges, although we note calibration improves for CIFAR100-C when augmenting every image.

For \am we additionally compare mixing augmented and un-augmented images with either a Bernoulli (Bern.) or Beta distribution (see Section~\ref{sec:am}). We found the simple baseline of `no mixing' ($p\!=\!1.0$ in Table~\ref{tab:aug}) often outperformed the other methods, especially in corrupted error rate where it always performed best, but even on i.i.d.\ data improves over a Beta distribution for CIFAR-10. However adding a small probability of using un-augmented data ($p\!=\!0.875$ in Table~\ref{tab:aug}) improved calibration on i.i.d.\ data, which matches the intuition that exposing the model to some entirely in-distribution images at training time will help calibrate the model on i.i.d.\ held-out data.

Finally, we combine adversarial perturbations and \am in the same model. This combination performed best on CIFAR10. When comparing on the test set, using adversarial perturbations performs the best on all metrics. We also outperform the best models (when using \am with the Jensen-Shannon divergence described in Section~\ref{sec:am}) of \citet{hendrycks*2020augmix}, who achieve 10.9\% error and 34.9\% error on CIFAR10-C and CIFAR100-C respectively.

\section{Conclusion}

We present a simple method for increasing calibration in ensembles: using a different data augmentation for each ensemble member. We additionally avoid problems with miscalibration on i.i.d.\ data by randomly mixing in some un-augmented images. Overall our methods improve both calibration and accuracy across i.i.d.\ and out-of-distribution data. In future we hope to add theoretical understanding to our simple principles, perhaps automatically determining which augmentation to use. \am is well suited to image data, and we hope to expand to other domains and datasets, especially those with discrete inputs where the perturbations described here do not obviously apply.

\section*{Acknowledgements}
We thank Artur Bekasov, Conor Durkan and James Ritchie for valuable discussion. 
Asa Cooper Stickland was supported in part by the EPSRC Centre for Doctoral Training in
Data Science, funded by the UK Engineering and Physical Sciences Research Council (grant EP/L016427/1) and the University of Edinburgh.

\bibliography{example_paper}

\begin{thebibliography}{22}
\providecommand{\natexlab}[1]{#1}
\providecommand{\url}[1]{\texttt{#1}}
\expandafter\ifx\csname urlstyle\endcsname\relax
  \providecommand{\doi}[1]{doi: #1}\else
  \providecommand{\doi}{doi: \begingroup \urlstyle{rm}\Url}\fi

\bibitem[Cubuk et~al.(2019)Cubuk, Zoph, Mane, Vasudevan, and
  Le]{Cubuk_2019_CVPR}
Cubuk, E.~D., Zoph, B., Mane, D., Vasudevan, V., and Le, Q.~V.
\newblock Autoaugment: Learning augmentation strategies from data.
\newblock In \emph{The IEEE Conference on Computer Vision and Pattern
  Recognition (CVPR)}, June 2019.

\bibitem[Fan et~al.(2020)Fan, Grave, and Joulin]{Fan2020Reducing}
Fan, A., Grave, E., and Joulin, A.
\newblock Reducing transformer depth on demand with structured dropout.
\newblock In \emph{International Conference on Learning Representations}, 2020.

\bibitem[Goodfellow et~al.(2015)Goodfellow, Shlens, and
  Szegedy]{goodfellow2014explaining}
Goodfellow, I., Shlens, J., and Szegedy, C.
\newblock Explaining and harnessing adversarial examples.
\newblock In \emph{International Conference on Learning Representations}, 2015.

\bibitem[Guo et~al.(2017)Guo, Pleiss, Sun, and Weinberger]{pmlr-v70-guo17a}
Guo, C., Pleiss, G., Sun, Y., and Weinberger, K.~Q.
\newblock On calibration of modern neural networks.
\newblock In Precup, D. and Teh, Y.~W. (eds.), \emph{Proceedings of the 34th
  International Conference on Machine Learning}, volume~70 of \emph{Proceedings
  of Machine Learning Research}, pp.\  1321--1330, International Convention
  Centre, Sydney, Australia, 06--11 Aug 2017. PMLR.

\bibitem[{Hansen} \& {Salamon}(1990){Hansen} and {Salamon}]{58871}
{Hansen}, L.~K. and {Salamon}, P.
\newblock Neural network ensembles.
\newblock \emph{IEEE Transactions on Pattern Analysis and Machine
  Intelligence}, 12\penalty0 (10):\penalty0 993--1001, 1990.

\bibitem[Hendrycks \& Dietterich(2019)Hendrycks and
  Dietterich]{hendrycks2019robustness}
Hendrycks, D. and Dietterich, T.
\newblock Benchmarking neural network robustness to common corruptions and
  perturbations.
\newblock \emph{Proceedings of the International Conference on Learning
  Representations}, 2019.

\bibitem[Hendrycks et~al.(2020)Hendrycks, Mu, Cubuk, Zoph, Gilmer, and
  Lakshminarayanan]{hendrycks*2020augmix}
Hendrycks, D., Mu, N., Cubuk, E.~D., Zoph, B., Gilmer, J., and
  Lakshminarayanan, B.
\newblock Augmix: A simple method to improve robustness and uncertainty under
  data shift.
\newblock In \emph{International Conference on Learning Representations}, 2020.

\bibitem[Huang et~al.(2016)Huang, Sun, Liu, Sedra, and
  Weinberger]{Huang2016DeepNW}
Huang, G., Sun, Y., Liu, Z., Sedra, D., and Weinberger, K.~Q.
\newblock Deep networks with stochastic depth.
\newblock In \emph{ECCV}, 2016.

\bibitem[Kim et~al.(2018)Kim, Goyal, Chawla, Lee, and Kwon]{Kim_2018_ECCV}
Kim, W., Goyal, B., Chawla, K., Lee, J., and Kwon, K.
\newblock Attention-based ensemble for deep metric learning.
\newblock In \emph{The European Conference on Computer Vision (ECCV)},
  September 2018.

\bibitem[Krizhevsky(2009)]{cifar}
Krizhevsky, A.
\newblock Learning multiple layers of features from tiny images, 2009.
\newblock \url{http://www.cs.toronto.edu/~kriz/cifar.html}.

\bibitem[Lakshminarayanan et~al.(2017)Lakshminarayanan, Pritzel, and
  Blundell]{NIPS2017_7219}
Lakshminarayanan, B., Pritzel, A., and Blundell, C.
\newblock Simple and scalable predictive uncertainty estimation using deep
  ensembles.
\newblock In Guyon, I., Luxburg, U.~V., Bengio, S., Wallach, H., Fergus, R.,
  Vishwanathan, S., and Garnett, R. (eds.), \emph{Advances in Neural
  Information Processing Systems 30}, pp.\  6402--6413. Curran Associates,
  Inc., 2017.

\bibitem[Lee et~al.(2016)Lee, Purushwalkam Shiva~Prakash, Cogswell, Ranjan,
  Crandall, and Batra]{NIPS2016_6270}
Lee, S., Purushwalkam Shiva~Prakash, S., Cogswell, M., Ranjan, V., Crandall,
  D., and Batra, D.
\newblock Stochastic multiple choice learning for training diverse deep
  ensembles.
\newblock In Lee, D.~D., Sugiyama, M., Luxburg, U.~V., Guyon, I., and Garnett,
  R. (eds.), \emph{Advances in Neural Information Processing Systems 29}, pp.\
  2119--2127. Curran Associates, Inc., 2016.

\bibitem[Loshchilov \& Hutter(2017)Loshchilov and
  Hutter]{DBLP:conf/iclr/LoshchilovH17}
Loshchilov, I. and Hutter, F.
\newblock {SGDR:} stochastic gradient descent with warm restarts.
\newblock In \emph{5th International Conference on Learning Representations,
  {ICLR} 2017, Toulon, France, April 24-26, 2017, Conference Track
  Proceedings}. OpenReview.net, 2017.

\bibitem[Ovadia et~al.(2019)Ovadia, Fertig, Ren, Nado, Sculley, Nowozin,
  Dillon, Lakshminarayanan, and Snoek]{NIPS2019_9547}
Ovadia, Y., Fertig, E., Ren, J., Nado, Z., Sculley, D., Nowozin, S., Dillon,
  J., Lakshminarayanan, B., and Snoek, J.
\newblock Can you trust your model's uncertainty? evaluating predictive
  uncertainty under dataset shift.
\newblock In Wallach, H., Larochelle, H., Beygelzimer, A., d'Alch\'{e} Buc, F.,
  Fox, E., and Garnett, R. (eds.), \emph{Advances in Neural Information
  Processing Systems 32}, pp.\  13991--14002. Curran Associates, Inc., 2019.

\bibitem[Pang et~al.(2019)Pang, Xu, Du, Chen, and Zhu]{pmlr-v97-pang19a}
Pang, T., Xu, K., Du, C., Chen, N., and Zhu, J.
\newblock Improving adversarial robustness via promoting ensemble diversity.
\newblock In Chaudhuri, K. and Salakhutdinov, R. (eds.), \emph{Proceedings of
  the 36th International Conference on Machine Learning}, volume~97 of
  \emph{Proceedings of Machine Learning Research}, pp.\  4970--4979, Long
  Beach, California, USA, 09--15 Jun 2019. PMLR.

\bibitem[Sinha et~al.(2020)Sinha, Bharadhwaj, Goyal, Larochelle, Garg, and
  Shkurti]{sinha2020dibs}
Sinha, S., Bharadhwaj, H., Goyal, A., Larochelle, H., Garg, A., and Shkurti, F.
\newblock Dibs: Diversity inducing information bottleneck in model ensembles,
  2020.

\bibitem[Srivastava et~al.(2014)Srivastava, Hinton, Krizhevsky, Sutskever, and
  Salakhutdinov]{srivastava2014}
Srivastava, N., Hinton, G., Krizhevsky, A., Sutskever, I., and Salakhutdinov,
  R.
\newblock Dropout: a simple way to prevent neural networks from overfitting.
\newblock \emph{Journal of Machine Learning Research}, 15:\penalty0 1929--1958,
  2014.

\bibitem[Wen et~al.(2020)Wen, Tran, and Ba]{Wen2020BatchEnsemble:}
Wen, Y., Tran, D., and Ba, J.
\newblock {BatchEnsemble}: an alternative approach to efficient ensemble and
  lifelong learning.
\newblock In \emph{International Conference on Learning Representations}, 2020.

\bibitem[Xie et~al.(2019)Xie, Hovy, Luong, and
  Le]{DBLP:journals/corr/abs-1911-04252}
Xie, Q., Hovy, E.~H., Luong, M., and Le, Q.~V.
\newblock Self-training with noisy student improves imagenet classification.
\newblock \emph{CoRR}, abs/1911.04252, 2019.

\bibitem[{Xie} et~al.(2017){Xie}, {Girshick}, {Dollár}, {Tu}, and
  {He}]{resnext}
{Xie}, S., {Girshick}, R., {Dollár}, P., {Tu}, Z., and {He}, K.
\newblock Aggregated residual transformations for deep neural networks.
\newblock In \emph{2017 IEEE Conference on Computer Vision and Pattern
  Recognition (CVPR)}, pp.\  5987--5995, 2017.

\bibitem[Yun et~al.(2019)Yun, Han, Oh, Chun, Choe, and Yoo]{Yun2019CutMixRS}
Yun, S., Han, D., Oh, S.~J., Chun, S., Choe, J., and Yoo, Y.
\newblock {C}ut{M}ix: Regularization strategy to train strong classifiers with
  localizable features.
\newblock \emph{2019 IEEE/CVF International Conference on Computer Vision
  (ICCV)}, pp.\  6022--6031, 2019.

\bibitem[Zhang et~al.(2018)Zhang, Cisse, Dauphin, and
  Lopez-Paz]{zhang2018mixup}
Zhang, H., Cisse, M., Dauphin, Y.~N., and Lopez-Paz, D.
\newblock mixup: Beyond empirical risk minimization.
\newblock In \emph{International Conference on Learning Representations}, 2018.

\end{thebibliography}
\bibliographystyle{icml2019}

\appendix
\begin{table*}[bp]
\vspace*{-0.1in}
\caption{Comparing performance when using the \am  (referred to as `AM' in the table) method, with either a Beta distribution to mix augmented images, or a Bernoulli (`Bern.') distribution, with $p$ being the probability of augmenting an input image. `No B.E.' refers to using a single model. We highlight in bold the best result for each metric.} 
\label{tab:app-aug}
\vskip 0.15in
\begin{center}
\scalebox{0.9}{
\begin{small}
\begin{sc}
\begin{tabular}{lccccccr}
\toprule
 & Cifar 10 &  & & Cifar 100 & & & \\

Method & Val.\ err.  & Val. ECE  & Val. ECE-rms & Val.\ err.  & Val. ECE  & Val. ECE-rms \\

AM (no B.E.) Beta    & {\bf 4.04}  & 1.44 & 3.54 & 22.4 &{\bf  8.22} &{\bf  10.1} \\
AM (no B.E.) Bern.\ ($p=0.875$)   & 4.46  &{\bf  1.35} & 3.54 &{\bf  21.6} & 8.42 & 10.2 \\
AM (no B.E.) Bern.\ ($p=1.0$)  & 4.30 & 1.47 &{\bf  2.90} & 21.7 & 8.33 & 10.7 \\

\midrule
\midrule

 & Cifar 10-C &  & & Cifar 100-C & & & \\

 & Err.  & ECE  & ECE-rms & Err.  & ECE  & ECE-rms \\

AM (no B.E.) Beta    & 13.2  & {\bf 5.36}& {\bf 7.06} & 37.8 & 15.7 & 17.4 \\
AM (no B.E.) Bern.\ ($p=0.875$)  & 13.1 & 5.38 & 7.18 &{\bf  36.7 }& 16.1 & 18.0 \\
AM (no B.E.) Bern.\ ($p=1.0$)  &  {\bf 13.0} & 5.41 & 7.18 & {\bf 36.7} &{\bf  15.8} &{\bf  17.7} \\

\bottomrule
\end{tabular}
\end{sc}
\end{small}}
\end{center}
\vskip -0.1in
\end{table*}
\section{Results without BatchEnsemble}
\label{app}

We compare (Table~\ref{tab:app-aug}) performance with \am for single models (No BatchEnsemble). No clear trends emerge except that always augmenting ($p=1.0$) leads to better performance on CIFAR10-C and CIFAR100-C. Perhaps surprisingly, augmenting all input images does not hurt i.i.d. accuracy. 

\end{document}